\documentclass{article}

\usepackage[final, nonatbib]{nips_2018}
\usepackage[utf8]{inputenc} 
\usepackage[T1]{fontenc}   
\usepackage{hyperref}       
\usepackage{url}            
\usepackage{booktabs}       
\usepackage{amsfonts}       
\usepackage{nicefrac}       
\usepackage{microtype}  
\usepackage{graphicx}
\usepackage{algorithm}
\usepackage{algorithmic}
\usepackage{subfigure}
\usepackage{float}
\usepackage{cite}
\usepackage{amsmath}
\usepackage{comment}
\usepackage{caption}

\title{Grow and Prune Compact, Fast, and Accurate LSTMs}

\author{
  Xiaoliang Dai$^{*}$\\
  Princeton University\\
  \texttt{xdai@princeton.edu} \\
   \And
   Hongxu Yin$^{*}$ \\
   Princeton University\\
  \texttt{hongxuy@princeton.edu} \\
   \And
   Niraj K. Jha \\
   Princeton University\\
  \texttt{jha@princeton.edu} \\
}
\begin{document}

\maketitle

\begin{abstract}
Long short-term memory (LSTM) has been widely used for sequential data 
modeling. Researchers have increased LSTM depth by stacking LSTM cells to 
improve performance. This incurs model redundancy, increases run-time delay, 
and makes the LSTMs more prone to overfitting. To address these problems, we 
propose a hidden-layer LSTM (H-LSTM) that adds hidden layers to LSTM's 
original one-level non-linear control gates. H-LSTM increases accuracy 
while employing fewer external stacked layers, thus reducing the number
of parameters and run-time latency significantly. We employ
grow-and-prune (GP) training to iteratively adjust the hidden layers 
through gradient-based growth and magnitude-based pruning of connections. 
This learns both the weights and the compact architecture of H-LSTM control 
gates. We have GP-trained H-LSTMs for image captioning and speech recognition
applications. For the NeuralTalk architecture on the MSCOCO dataset, 
our three models reduce the number of parameters by 38.7$\times$ 
[floating-point operations (FLOPs) by 45.5$\times$], run-time latency by 
4.5$\times$, and improve the CIDEr score by 2.6. For the DeepSpeech2 
architecture on the AN4 dataset, our two models reduce the number of 
parameters by 19.4$\times$ (FLOPs by 23.5$\times$), run-time latency by 15.7\%, 
and the word error rate from 12.9\% to 8.7\%. Thus, GP-trained
H-LSTMs can be seen to be compact, fast, and accurate.

\end{abstract}

\section{Introduction}
\footnotetext{$^{*}$ indicates equal contribution. This work was supported by NSF Grant No. CNS-1617640.}
Recurrent neural networks (RNNs) have been ubiquitously employed for 
sequential data modeling because of their ability to carry information 
through recurrent cycles. Long short-term memory (LSTM) is a special
type of RNN that uses control gates and cell states to alleviate the 
exploding and vanishing gradient problems~\cite{lstm}. LSTMs are adept at 
modeling long-term and short-term dependencies. They deliver state-of-the-art 
performance for a wide variety of applications, such as speech 
recognition~\cite{deepspeech2}, image captioning~\cite{neuraltalk}, and 
neural machine translation~\cite{seq2seq}.

Researchers have kept increasing the model depth and size to improve the
performance of LSTMs. For example, the DeepSpeech2 architecture
\cite{deepspeech2}, which has been used for speech recognition, is more than 
2$\times$ deeper and 10$\times$ larger than the initial DeepSpeech
architecture proposed in~\cite{deepspeech1}.  However, large LSTM models may 
suffer from three problems.  First, deployment of a large LSTM model consumes 
substantial storage, memory bandwidth, and computational resources. Such 
demands may be too excessive for mobile phones and embedded devices. Second, 
large LSTMs are prone to overfitting but hard to 
regularize~\cite{lstmdropout}. Employing standard regularization 
methods that are used for feed-forward neural networks (NNs), such as 
dropout, in an LSTM cell is challenging~\cite{dropout,lstmdropout2}. Third, 
the increasingly stringent run-time latency constraints in real-time 
applications make large LSTMs, which incur high latency, inapplicable in
these scenarios.  All these problems pose a significant design challenge in 
obtaining compact, fast, and accurate LSTMs.

In this work, we tackle these design challenges simultaneously by
combining two novelties. We first propose a hidden-layer LSTM (H-LSTM) that 
introduces multi-level abstraction in the LSTM control gates by adding
several hidden layers. Then, we describe a grow-and-prune (GP) 
training method that combines gradient-based growth~\cite{nest} and 
magnitude-based pruning~\cite{ese} techniques to learn the weights and derive 
compact architectures for the control gates. This yields inference models that 
outperform the baselines from all three targeted design perspectives.

\section{Related work}
Going deeper is a common strategy for improving LSTM performance.  The 
conventional approach of stacking LSTM cells has shown significant
performance improvements on a wide range of applications, such as speech 
recognition and machine translation
\cite{deeplstm, deeplstm1, deeplstm2dropout}.  The recently proposed skipped 
connection technique has made it possible to train very deeply stacked LSTMs. 
This leads to high-performance architectures, such as residual 
LSTM~\cite{residuallstm} and highway LSTM~\cite{highwaylstm}.

Stacking LSTMs improves accuracy but incurs substantial computation and storage 
costs. Numerous recent approaches try to shrink the size of large NNs. A 
popular direction is to simplify the matrix representation. Typical techniques 
include matrix factorization~\cite{factorize}, low rank 
approximation~\cite{cnnlinear}, and basis filter set 
reduction~\cite{cnnlowrank}. Another direction focuses on efficient storage and 
representation of weights. Various techniques, such as weight sharing within 
Toeplitz matrices~\cite{toeplitz}, weight tying through effective 
hashing~\cite{cnnhash}, and appropriate weight 
quantization~\cite{quantization, quantization1, quantization2}, can greatly 
reduce model size, in some cases at the expense of a slight performance 
degradation.

Network pruning has emerged as another popular approach for LSTM compression. 
Han et al.~show that pruning can significantly cut down on the size of 
deep convolutional neural networks (CNNs) and LSTMs~\cite{PruningHS, ese, systprune}.  
Moreover, Yu et al.~show that post-pruning sparsity in weight matrices 
can even improve 
speech recognition accuracy~\cite{sparselstm}. Narang el
al.~incorporate pruning in the training process and compress the LSTM 
model size by approximately 10$\times$ while reducing training time 
significantly~\cite{baiduprune}. 

Apart from size reduction, run-time latency reduction has also attracted an 
increasing amount of research attention.  A recent work by Zen et al.~uses 
unidirectional LSTMs with a recurrent output layer to reduce run-time 
latency~\cite{unidirectional}. Amodei et al.~propose a more efficient beam 
search strategy to speed up inference with only a minor accuracy 
loss~\cite{deepspeech2}. Narang et al.~exploit hardware-driven cuSPARSE 
libraries on a TitanX Maxwell GPU to speed up post-pruning sparse LSTMs by 
1.16$\times$ to 6.80$\times$~\cite{baiduprune}. 

\section{Methodology}

In this section, we explain our LSTM synthesis methodology that is based
on H-LSTM cells and GP training. We first describe the H-LSTM structure, after 
which we illustrate GP training in detail.

\subsection{Hidden-layer LSTM}

Recent years have witnessed the impact of increasing NN depth on its 
performance.  A deep architecture allows an NN to capture low/mid/high-level 
features through a multi-level abstraction. However, since a conventional 
LSTM employs fixed single-layer non-linearity for gate controls, the current 
standard approach for increasing model depth is through stacking several 
LSTM cells externally. In this work, we argue for a different approach that 
increases depth within LSTM cells. We propose an H-LSTM whose control gates are 
enhanced by adding hidden layers. We show that stacking fewer H-LSTM cells 
can achieve higher accuracy with fewer parameters and smaller run-time 
latency than conventionally stacked LSTM cells.

\begin{figure*}[t]
\begin{center}
\includegraphics[width=135mm]{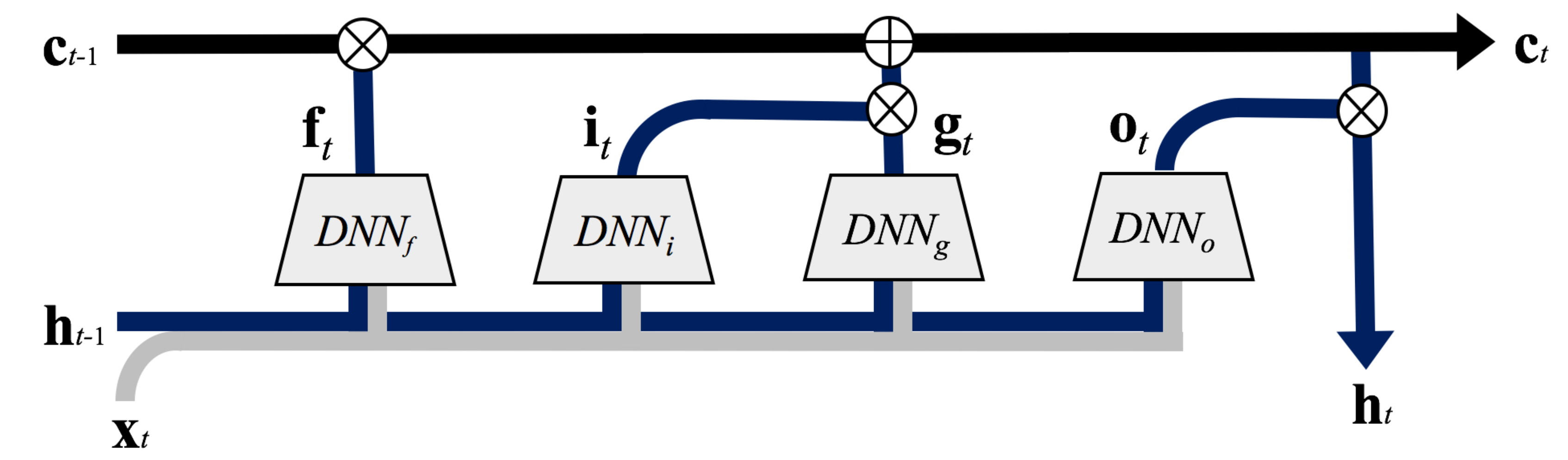}
\end{center}
\caption{Schematic diagram of H-LSTM.}
\label{fig:main_cell}
\end{figure*}

We show the schematic diagram of an H-LSTM in Fig.~\ref{fig:main_cell}. The 
internal computation flow is governed by Eq.~(\ref{eq:main_cell}), where 
$\textbf{f}_{t}$, $\textbf{i}_{t}$, $\textbf{o}_{t}$, $\textbf{g}_{t}$, 
$\textbf{x}_{t}$, $\textbf{h}_{t}$, and $\textbf{c}_{t}$ refer to the forget 
gate, input gate, output gate, vector for cell updates, input, hidden state, 
and cell state at step $t$, respectively; $\textbf{h}_{t-1}$ and 
$\textbf{c}_{t-1}$ refer to the hidden and cell states at step $t-1$; 
$DNN$, $H$, $\textbf{W}$, $\textbf{b}$, $\sigma$, and $\otimes$ refer to the 
deep neural network (DNN) gates, hidden layers (each performs a linear 
transformation followed by the activation function), weight matrix, bias, 
$sigmoid$ function, and element-wise multiplication, respectively; $^{*}$ 
indicates zero or more $H$ layers for the DNN gate.

\begin{equation}
\begin{split}
 \left(
 \begin{matrix}
   \textbf{f}_{t}  \\
   \textbf{i}_{t}  \\
   \textbf{o}_{t}  \\
   \textbf{g}_{t}  
  \end{matrix}
  \right) 
  &=
   \left(
 \begin{matrix}
   DNN_{f}([\textbf{x}_{t},\textbf{h}_{t-1}])  \\
   DNN_{i}([\textbf{x}_{t},\textbf{h}_{t-1}])  \\
   DNN_{o}([\textbf{x}_{t},\textbf{h}_{t-1}])  \\
   DNN_{g}([\textbf{x}_{t},\textbf{h}_{t-1}])  
  \end{matrix}
  \right) 
  =
 \left(
 \begin{matrix}
   \sigma(\textbf{W}_f H^{*}( [\textbf{x}_{t},\textbf{h}_{t-1}])+\textbf{b}_f) \\
   \sigma(\textbf{W}_i H^{*}( [\textbf{x}_{t},\textbf{h}_{t-1}])+\textbf{b}_i) \\
   \sigma(\textbf{W}_o H^{*}( [\textbf{x}_{t},\textbf{h}_{t-1}])+\textbf{b}_o) \\
   tanh(\textbf{W}_g H^{*}( [\textbf{x}_{t},\textbf{h}_{t-1}])+\textbf{b}_g)
  \end{matrix}
  \right)  \\
  &\ \ \ \ \ \ \ \ \ \ \ \ \ \ \ \ \ \ \ \ \ \ \ \ \ \ \ \  \textbf{c}_{t} = \textbf{f}_{t} \otimes \textbf{c}_{t-1} + \textbf{i}_{t} \otimes \textbf{g}_{t}\\
  &\ \ \ \ \ \ \ \ \ \ \ \ \ \ \ \ \ \ \ \ \ \ \ \ \ \ \ \ \ \ \ \  \textbf{h}_{t} = \textbf{o}_{t} \otimes tanh(\textbf{c}_t)
\end{split}
\label{eq:main_cell}
\end{equation}

The introduction of DNN gates provides three major benefits to an H-LSTM:
\begin{enumerate}
\item \textbf{Strengthened control}: Hidden layers in DNN gates enhance gate 
control through multi-level abstraction. This makes an H-LSTM more capable 
and intelligent, and alleviates its reliance on external stacking. 
Consequently, an H-LSTM can achieve comparable or even improved accuracy with 
fewer external stacked layers relative to a conventional LSTM, leading to 
higher compactness.
\item \textbf{Easy regularization}: The conventional approach only uses
dropout in the input/output layers and recurrent connections in the LSTMs.  
In our case, it becomes possible to apply dropout even to all control gates 
within an LSTM cell. This reduces overfitting and leads to better 
generalization.
\item \textbf{Flexible gates}: Unlike the fixed but specially-crafted gate 
control functions in LSTMs, DNN gates in 
an H-LSTM offer a wide range of choices for internal activation
functions, such as the ReLU. This may provide additional benefits to the model. 
For example, networks typically learn faster with ReLUs~\cite{dropout}. They can also take advantage of ReLU's zero outputs for FLOPs reduction.

\end{enumerate}

\subsection{Grow-and-Prune Training}

Conventional training based on back propagation on fully-connected NNs yields 
over-parameterized models. Han et al.~have successfully implemented pruning 
to drastically reduce the size of large CNNs and LSTMs~\cite{PruningHS, ese}. 
The pruning phase is complemented with a brain-inspired growth phase for large 
CNNs in~\cite{nest}. The network growth phase allows a CNN to grow neurons, 
connections, and feature maps, as necessary, during training. Thus, it enables 
automated search in the architecture space. It has been shown that a sequential 
combination of growth and pruning can yield additional compression on CNNs 
relative to pruning-only methods (e.g., 1.7$\times$ for AlexNet and 
2.0$\times$ for VGG-16 on top of the pruning-only methods)~\cite{nest}. 
In this work, we extend GP training to LSTMs.

\begin{figure*}[h]
\begin{center}
\includegraphics[width=110mm]{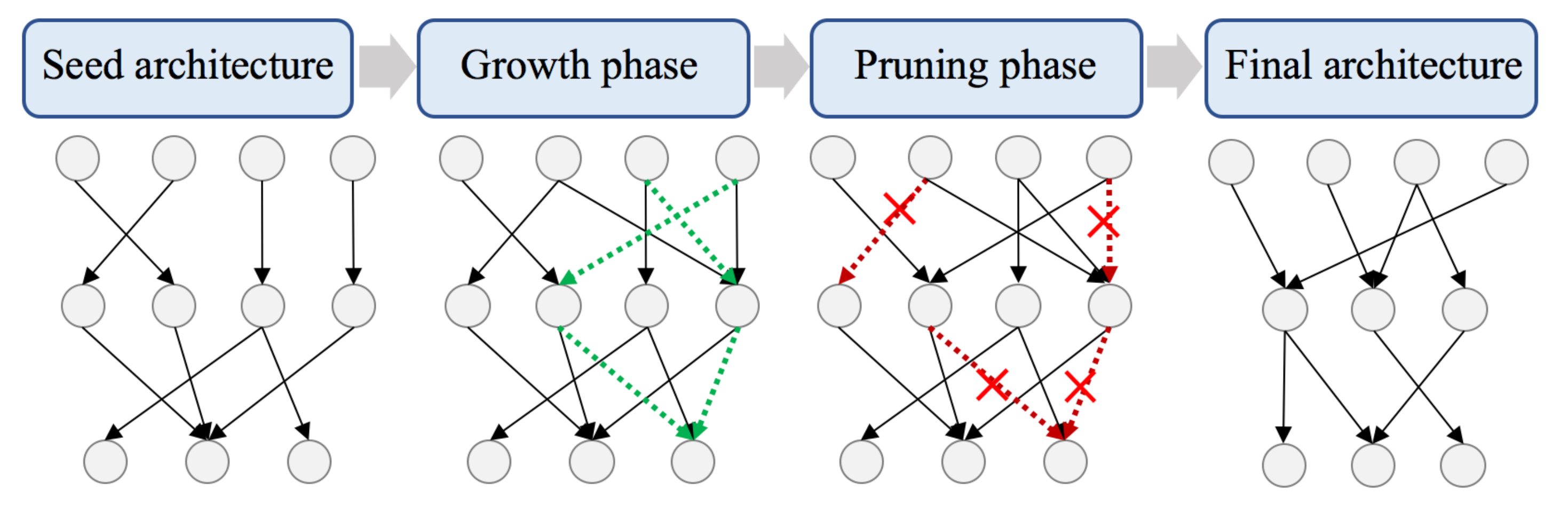}
\end{center}
\caption{An illustration of GP training.}
\label{fig:phases}
\end{figure*}

We illustrate the GP training flow in Fig.~\ref{fig:phases}. It starts from 
a randomly initialized sparse seed architecture. The seed architecture contains 
a very limited fraction of connections to facilitate initial gradient 
back-propagation. The remaining connections in the matrices are dormant and 
masked to zero.  The flow ensures that all neurons in the network are 
connected.  During training, it first grows connections based on the gradient 
information.  Then, it prunes away redundant connections for compactness, based 
on their magnitudes. Finally, GP training rests at an accurate, yet compact, 
inference model. We explain the details of each phase next.

GP training adopts the following policies:

\noindent
\textbf{Growth policy:} Activate a dormant $w$ in \textbf{W} iff $|w.grad|$ 
is larger than the $(100\alpha)^{th}$ percentile of all elements in 
|\textbf{W}$.grad$|.

\textbf{Pruning policy:} Remove a $w$ iff $|w|$ is smaller than 
the $(100\beta)^{th}$ percentile of all elements in |\textbf{W}|.

$w$, \textbf{W}, $.grad$, $\alpha$, and $\beta$ refer to the weight of a 
single connection, weights of all connections within one layer, operation 
to extract the gradient, growth ratio, and pruning ratio, respectively.

In the growth phase, the main objective is to locate the most effective dormant 
connections to reduce the value of loss function $L$. We first evaluate 
$\partial L / \partial w$ for each dormant connection $w$ based on its average 
gradient over the entire training set. Then, we activate each dormant 
connection whose gradient magnitude $|w.grad| = |\partial L / \partial w|$ 
surpasses the $(100\alpha)^{th}$ percentile of the gradient magnitudes
of its corresponding weight matrix. This rule caters to dormant connections 
iff they provide most efficiency in $L$ reduction. Growth can also help avoid 
local minima, as observed by Han et al.~in their dense-sparse-dense training 
algorithm to improve accuracy~\cite{dsd}. 

We prune the network after the growth phase. Pruning of insignificant weights 
is an iterative process. In each iteration, we first prune away insignificant 
weights whose magnitudes are smaller than the $(100\beta)^{th}$ percentile 
within their respective layers. We prune a neuron if all its input (or output) 
connections are pruned away. We then retrain the NN after weight pruning to 
recover its performance before starting the next pruning iteration.  The 
pruning phase terminates when retraining cannot achieve a pre-defined accuracy 
threshold. GP training finalizes a model based on the last complete iteration. 

\section{Experimental Results}

We present our experimental results for image captioning and speech recognition 
benchmarks next.  We implement our experiments using PyTorch~\cite{pytorch} on 
Nvidia GTX 1060 and Tesla P100 GPUs. 

\subsection{NeuralTalk for Image Captioning}
We first show the effectiveness of our proposed methodology 
on image captioning.  

\textbf{Architecture:} 
We experiment with 
the NeuralTalk architecture~\cite{neuraltalk2,neuraltalk} that uses the 
last hidden layer of a pre-trained CNN image encoder as an input to a 
recurrent decoder for sentence generation. The recurrent decoder applies a 
beam search technique for sentence generation. A beam size of $k$ indicates 
that at step $t$, the decoder considers the set of $k$ best sentences
obtained so far as candidates to generate sentences in step $t + 1$, and 
keeps the best $k$ results~\cite{neuraltalk2, neuraltalk,showtell_google}. 
In our experiments, we use VGG-16~\cite{vgg} as the CNN encoder, same as 
in~\cite{neuraltalk2, neuraltalk}. We then use H-LSTM and LSTM
cells with the same width of 512 for the recurrent decoder and compare their 
performance. We use $Beam=2$ as the default beam size.

\textbf{Dataset:} We report results on the MSCOCO dataset~\cite{mscoco}.  It 
contains 123287 images of size 256$\times$256$\times$3, along with five 
reference sentences per image.  We use the publicly available 
split~\cite{neuraltalk2}, which has 113287, 5000, and 5000 images in the 
training, validation, and test sets, respectively.

\textbf{Training:} We use the Adam optimizer~\cite{Adam} for this experiment. 
We use a batch size of 64 for training. We initialize the learning rate at 
$3$$\times$$ 10^{-4}$. In the first 90 epochs, we fix the weights of the CNN 
and train the LSTM decoder only. We decay the learning rate by 0.8 every
six epochs 
in this phase. After 90 epochs, we start to fine-tune both the CNN and LSTM 
at a fixed $1$$\times$$10^{-6}$ learning rate.  We use a dropout ratio of 0.2 
for the hidden layers in the H-LSTM.  We also use a dropout ratio of 0.5 for 
the input and output layers of the LSTM, same as in~\cite{lstmdropout}.  We 
use CIDEr score~\cite{cider} as our evaluation criterion.

\subsubsection{Cell Comparison}
We first compare the performance of a fully-connected H-LSTM with a 
fully-connected LSTM to show the benefits emanating from using the H-LSTM 
cell alone. 

\begin{figure}
\begin{minipage}{66mm}
\centering
\includegraphics[width=66mm]{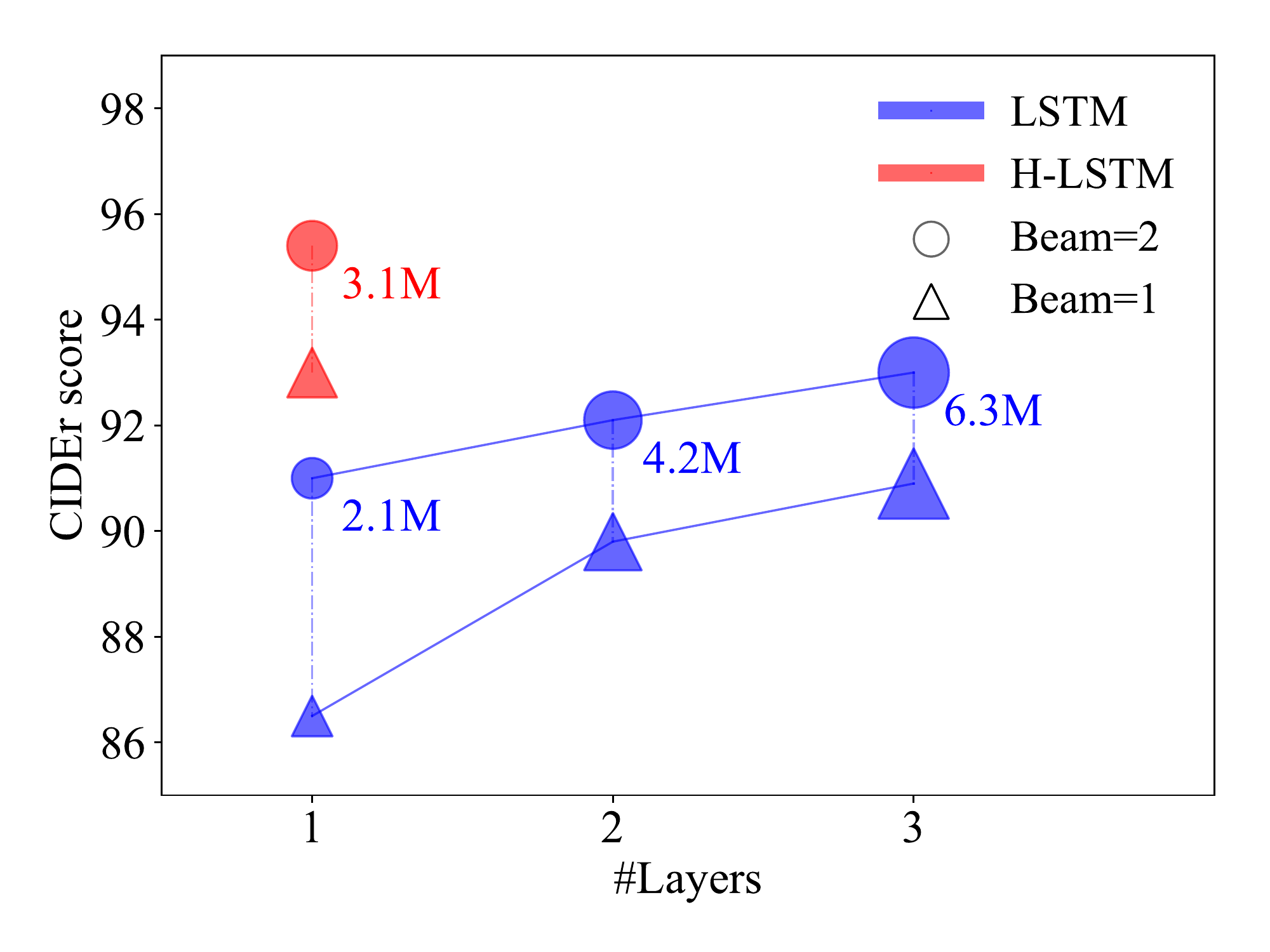}
\caption{Comparison of NeuralTalk CIDEr for the LSTM and H-LSTM cells. Number 
and area indicate size.}
\label{fig:nt_cell}
\end{minipage}
\begin{minipage}{74mm}
\small
\vspace{-12mm}
\captionof{table}{Cell comparison for the NeuralTalk architecture on the 
MSCOCO dataset.}
\small
\begin{tabular}{lccc}
\hline
Model &  CIDEr & \#Param & Latency\\
\hline
Single LSTM   & 91.0 & 2.1M & 21ms  \\
Stacked 2-layer LSTMs   & 92.1 & 4.2M & 29ms\\
Stacked 3-layer LSTMs   & 92.8 & 6.3M & 36ms\\
\hline
\textbf{H-LSTM (Beam=2)}  & \textbf{95.4} & \textbf{3.1M} & \textbf{24ms} \\
\textbf{H-LSTM (Beam=1)}  & \textbf{93.0} & \textbf{3.1M} & \textbf{8ms}\\
\hline
\label{tab:Neuraltalk}
\end{tabular}
\end{minipage}
\end{figure}

The NeuralTalk architecture with a single LSTM achieves a 91.0 CIDEr 
score~\cite{neuraltalk}.  We also experiment with stacked 2-layer and 3-layer 
LSTMs, which achieve 92.1 and 92.8 CIDEr scores, respectively.  We next train 
a single H-LSTM and compare the results in Fig.~\ref{fig:nt_cell} and 
Table~\ref{tab:Neuraltalk}.  
Our single H-LSTM achieves a CIDEr score of 95.4, which is 4.4, 3.3, 2.6 
higher than the single LSTM, stacked 2-layer LSTM, and stacked 3-layer LSTM, 
respectively.

H-LSTM can also reduce run-time latency.  Even with $Beam=1$, a single H-LSTM 
achieves a higher accuracy than the three LSTM baselines.  Reducing the beam 
size leads to run-time latency reduction.  H-LSTM is 4.5$\times$, 3.6$\times$, 
2.6$\times$ faster than the stacked 3-layer LSTM, stacked 2-layer LSTM, 
and single LSTM, respectively, while providing higher accuracy.

\subsubsection{Training Comparison}
Next, we implement both network pruning and our GP training to synthesize 
compact inference models for an H-LSTM ($Beam = 2$).  The seed architecture for GP training has a sparsity 
of 50\%. In the growth phase, we use a 0.8 growth ratio in the first
five epochs.  We summarize the results in Table~\ref{tab:neuraltalk_methods}, 
where CR refer to the compression ratio relative to a fully-connected model.  
GP training provides an additional 1.40$\times$ improvement on CR 
compared with network pruning.

\begin{table}[h]
\centering
\small
\caption{Training algorithm comparison}
\begin{tabular}{lcccccc}
\hline
Method & Cell & \#Layers & CIDEr & \#Param & CR & Improvement \\
\hline
Network pruning  & H-LSTM & 1 & 95.4 & 550K & 5.7$\times$ & - \\
\textbf{GP training} & \textbf{H-LSTM} & \textbf{1} & \textbf{95.4} & \textbf{394K} &\textbf{8.0$\times$} &\textbf{1.40$\times$} \\
\hline
\label{tab:neuraltalk_methods}
\end{tabular}
\end{table}

\subsubsection{Inference Model Comparison}
We list our GP-trained H-LSTM models in Table~\ref{tab:neuraltalk_results}.  Note 
that the accurate and fast models are the same network with different beam 
sizes. The compact model is obtained through further pruning of the accurate 
model.  We choose the stacked 3-layer LSTM as our baseline due to its high 
accuracy.  Our accurate, fast, and compact models demonstrate improvements in 
all aspects (accuracy, speed, and compactness), with a 2.6 higher CIDEr score, 
4.5$\times$ speedup, and 38.7$\times$ fewer parameters, respectively.

\begin{table}
\centering
\small
\caption{Different inference models for the MSCOCO dataset}
\begin{tabular}{lccccccc}
\hline
Model & \#Layers & Cell & Beam & CIDEr & \#Param & FLOPs & Latency \\
\hline
Single LSTM & 1 & LSTM & 2 & 91.0 & 2.1M & 4.2M & 21ms \\
Stacked LSTMs  & 2 & LSTM & 2 & 92.1 & 4.2M & 8.4M &  29ms \\
Stacked LSTMs  & 3 & LSTM & 2 & 92.8 & 6.3M & 12.6M &  36ms \\
\hline
\textbf{Ours: accurate} & 1 & H-LSTM & 2 & 95.4 & 394K & 670K & 24ms\\
\textbf{Ours: fast} & 1 & H-LSTM & 1 & 93.0 & 394K & 670K &  8ms \\
\textbf{Ours: compact} & 1 & H-LSTM & 2 & 93.3 & 163K & 277K & 24ms \\
\hline
\label{tab:neuraltalk_results}
\end{tabular}
\end{table}

\subsection{DeepSpeech2 for Speech Recognition}

We now consider another well-known application: speech recognition.

\textbf{Architecture:} We implement a bidirectional DeepSpeech2 architecture 
that employs stacked recurrent layers following convolutional layers for 
speech recognition~\cite{deepspeech2}. We use Mel-frequency cepstral 
coefficients (MFCCs) as network inputs, extracted from raw speech data at 
a 16KHz sampling rate and 20ms feature extraction window. There are two CNN 
layers prior to the recurrent layers and one connectionist temporal 
classification layer for decoding~\cite{ctc} after the recurrent layers. 
The width of the hidden and cell states is 800, same as 
in~\cite{stanford, github}. We also set the width of H-LSTM hidden layers 
to 800. 

\textbf{Dataset:} We use the AN4 dataset~\cite{an4} to evaluate the performance 
of our DeepSpeech2 architecture. It contains 948 training utterances and 130 
testing utterances. 

\textbf{Training:} We utilize a Nesterov SGD optimizer in our experiment. We 
initialize the learning rate to $3$$\times$$ 10^{-4}$, decayed per epoch by 
0.99. We use a batch size of 16 for training. We use a dropout ratio of 
0.2 for the hidden layers in the H-LSTM. We apply batch normalization between 
recurrent layers.  We apply L2 regularization during training with a weight 
decay of $1$$\times$$10^{-4}$.  We use word error rate (WER) as our evaluation 
criterion, same as in \cite{stanford,ethmit,github}.

\subsubsection{Cell Comparison}

We first compare the performance of the fully-connected H-LSTM against 
the fully-connected LSTM and gate recurrent unit (GRU) to 
demonstrate the benefits provided by the H-LSTM cell alone. GRU uses
reset and update gates for memory control and has fewer parameters than LSTM~\cite{gru}.

For the baseline, we train various DeepSpeech2 models containing a different 
number of stacked layers based on GRU and LSTM cells. The stacked 4-layer and 
5-layer GRUs achieve a WER of 14.35\% and 11.60\%, respectively.  The stacked 
4-layer and 5-layer LSTMs achieve a WER of 13.99\% and 10.56\%, respectively.

We next train an H-LSTM to make a comparison. Since an H-LSTM is intrinsically 
deeper, we aim to achieve a similar accuracy with a smaller stack. We reach 
a WER of 12.44\% and 8.92\% with stacked 2-layer and 3-layer H-LSTMs, 
respectively.

\begin{figure}
\begin{minipage}{72mm}
\centering
\includegraphics[width=72mm]{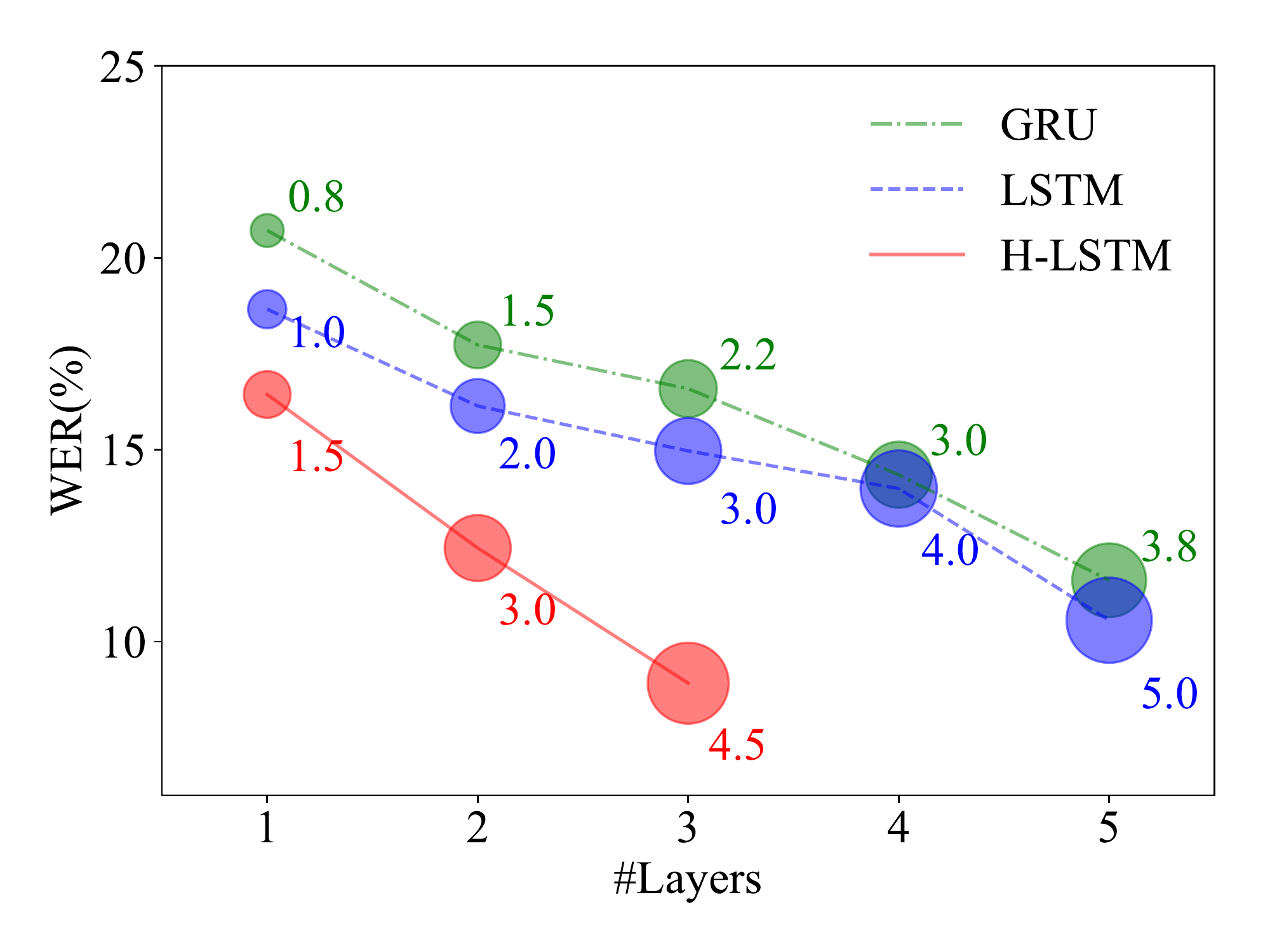}
\small
\caption{Comparison of DeepSpeech2 WERs for the GRU, LSTM, and H-LSTM cells. 
Number and area indicate relative size to one LSTM.}
\label{fig:ds_cell_fig}
\end{minipage}
\begin{minipage}{12mm}
\end{minipage}
\begin{minipage}{65mm}
\centering
\vspace{-8mm}
\captionof{table}{Cell comparison for the DeepSpeech2 architecture on
the AN4 dataset}
\small
\begin{tabular}{lccc}
\hline
Cell Type & \#Layers & Size & WER \\
\hline
GRU     & 4 & 3.0 & 14.35\%  \\
LSTM    & 4 & 4.0 & 13.99\%  \\
\textbf{H-LSTM} & \textbf{2} & \textbf{3.0} & \textbf{12.44\%} \\
\hline
GRU     & 5 & 3.8 & 11.64\%  \\
LSTM    & 5 & 5.0 & 10.56\%   \\
\textbf{H-LSTM} & \textbf{3} & \textbf{4.5} & \textbf{8.92\%} \\
\hline
\label{tb:ds_cell_table}
\end{tabular}
\end{minipage}
\end{figure}

We summarize the cell comparison results in Fig.~\ref{fig:ds_cell_fig} and 
Table~\ref{tb:ds_cell_table}, where all the sizes are normalized to the size of 
a single LSTM. We can see that H-LSTM can reduce WER by more than 1.5\% with 
two fewer layers relative to LSTMs and GRUs, thus satisfying our initial design 
goal to stack fewer cells that are individually deeper. H-LSTM models contain 
fewer parameters for a given target WER, and can achieve lower WER for a 
given number of parameters.

\subsubsection{Training Comparison}
We next implement GP training to show its additional benefits on top of 
just performing network pruning. We select the stacked 3-layer H-LSTMs for 
this experiment due to its highest accuracy. For GP training, we initialize 
the seed architecture with a connection sparsity of 50\%. We grow the networks 
for three epochs using a 0.9 growth ratio.

For best compactness, we set an accuracy threshold for both GP training and 
the pruning-only process at 10.52\% (lowest WER from relevant work 
\cite{stanford, ethmit, github}). We compare these two approaches in 
Table~\ref{tb:ds_methods}.  Compared to network pruning, GP training can 
further boost the CR by 2.44$\times$ while improving the accuracy slightly. 
This is consistent with prior observations that pruning large CNNs 
potentially inherits certain redundancies from the original fully-connected 
model that the growth phase can alleviate~\cite{nest}.

\begin{table}[t]
\centering
\small
\caption{Training algorithm comparison}
\begin{tabular}{lcccccc}
\hline
Method & Cell & \#Layers & WER & \#Param & CR &  Improvement \\
\hline
Network pruning & H-LSTM & 3 & 10.49\% &  6.4M & 6.8$\times$ & - \\
\textbf{GP training}    & \textbf{H-LSTM} & \textbf{3} & \textbf{10.37\%} & \textbf{2.6M} & \textbf{17.2$\times$} & \textbf{2.44$\times$} \\
\hline
\label{tb:ds_methods}
\end{tabular}
\end{table}

\subsubsection{Inference Model Comparison}
We obtain two GP-trained models by varying the WER constraint during the 
pruning phase: an accurate model aimed at a higher accuracy (9.00\% WER 
constraint), and a compact model aimed at extreme compactness (10.52\% WER 
constraint). 
 
We compare our results against prior work from the literature in 
Table~\ref{tb:ds_results}. We select a stacked 5-layer LSTM~\cite{stanford} 
as our baseline.
On top of the substantial parameter and FLOPs reductions, both the accurate and 
compact models also reduce the average run-time latency per instance from 
691.4ms to 583.0ms (15.7\% reduction) even without any sparse matrix library 
support.
 
\begin{table}[h]
\centering \small
\caption{Different inference models for the AN4 dataset}
\begin{tabular}{lccccc}
\hline
Model & RNN Type & WER(\%) & $\Delta$WER(\%)& \#Param(M) & FLOPs(M) \\
\hline
Lin et al.~\cite{stanford}       & LSTM  & 12.90 & -     & 50.4 (1.0$\times$) & 100.8 (1.0$\times$) \\
Alistarh et al.~\cite{ethmit}    & LSTM  & 18.85 & +5.95 & 13.0 (3.9$\times$) & 26.0 (3.9$\times$) \\
Sharen et al.~\cite{github}      & GRU   & 10.52 & -2.38 & 37.8 (1.3$\times$) & 75.7 (1.3$\times$) \\
\hline
\textbf{Ours: accurate} & \textbf{H-LSTM} & \textbf{8.73} & \textbf{-4.17} & \textbf{22.5 (2.3$\times$)} & \textbf{37.1 (2.9$\times$)} \\
\textbf{Ours: compact} & \textbf{H-LSTM}  & \textbf{10.37} & \textbf{-2.53} & \textbf{2.6 (19.4$\times$)} & \textbf{4.3 (23.5$\times$)} \\
\hline
\label{tb:ds_results}
\end{tabular}
\end{table}

The introduction of the ReLU activation function in DNN gates provides 
additional FLOPs reduction for the H-LSTM. This effect does not apply to 
LSTMs and GRUs that only use $tanh$ and $sigmoid$ gate control functions. 
At inference time, the average activation percentage of the ReLU outputs is
48.3\% for forward-direction LSTMs, and 48.1\% for backward-direction LSTMs. 
This further reduces the overall run-time FLOPs by 14.5\%. 

\begin{table}[h]
\centering \small
\caption{GP-trained compact 3-layer H-LSTM DeepSpeech2 model at 10.37\% WER}
\begin{tabular}{l|c|c|c}
\hline
 & Sparsity & Sparsity & Sparsity \\
\hline
Layers & Seed & Post-Growth & {Post-Pruning}  \\
\hline
H-LSTM layer1 & 50.00\% & 38.35\% & 94.26\% \\
H-LSTM layer2 & 50.00\% & 37.68\% & 94.20\% \\
H-LSTM layer3 & 50.00\% & 37.86\% & 94.21\% \\
\hline
Total & 50.00\% & 37.96\% & 94.22\% \\
\hline
\end{tabular}
\label{tb:ds_final_model}
\end{table}

The details of the final inference models are summarized in 
Table~\ref{tb:ds_final_model}. The final sparsity of the compact model is 
as high as 94.22\% due to the compounding effect of growth and pruning. 

\section{Discussions}
We observe the importance of regularization in H-LSTM on its final performance. 
We summarize the comparison between fully-connected models with and without 
dropout for both applications in Table~\ref{tb:dropout}. By
appropriately regularizing DNN gates, we improve the CIDEr score by 2.0 on 
NeuralTalk and reduce the WER from 9.88\% to 8.92\% on DeepSpeech2.

\begin{table}[h]
\centering \small
\caption{Impact of dropout on H-LSTM}
\begin{tabular}{lcc|lcc}
\hline
Architecture & Dropout & CIDEr & Architecture & Dropout & WER \\
\hline
NeuralTalk   & N   & 93.4 & DeepSpeech2 & N & 9.88\% \\
\hline
NeuralTalk   & Y   & 95.4 & DeepSpeech2 & Y & 8.92\% \\
\hline
\label{tb:dropout}
\end{tabular}
\end{table}

Some real-time applications may emphasize stringent memory and delay 
constraints instead of accuracy. In this case, the deployment of stacked 
LSTMs may be infeasible. The extra parameters used in H-LSTM's hidden layers 
may also seem disadvantageous in this scenario.  However, we next show that 
the extra parameters can be easily compensated by a reduced hidden layer 
width.  We compare several models for image captioning in 
Table~\ref{tab:slim_hlstm}. If we reduce the width of the hidden layers and 
cell states in the H-LSTM from 512 to 320, we can easily arrive at a 
single-layer H-LSTM that dominates the conventional LSTM from all three design 
perspectives. Our observation coincides with prior experience with neural 
network training where slimmer but deeper NNs (in this case H-LSTM) normally 
exhibit better performance than shallower but wider NNs (in this case LSTM). 

\begin{table}[h]
\centering \small
\caption{H-LSTM with reduced width for further speedup and compactness}
\begin{tabular}{lccc|cc|cc}
\hline
Cell & \#Layers & Width & \#Param  & CIDEr & Latency & CIDEr & Latency\\
\hline
& & & & \multicolumn{2}{c|}{Beam=2}& \multicolumn{2}{c}{Beam=1}\\
\hline
LSTM & 1 & 512 & 2.1M & 91.0 & 21ms & 86.5 & 7ms \\
H-LSTM & 1 & 512 & 3.1M & 95.4 & 24ms & 93.0 & 8ms \\
\hline
H-LSTM & 1 &  320 &  1.5M & 92.2 & 18ms & 88.1 & 5ms \\
\hline
\label{tab:slim_hlstm}
\end{tabular}
\end{table}

We also employ an activation function shift trick in our experiment. In the 
growth phase, we adopt a leaky ReLU (reverse slope of 0.01) as the activation 
function for $H^{*}$ in Eq.~(\ref{eq:main_cell}). Leaky ReLU effectively 
alleviates the `dying ReLU' 
phenomenon, in which a zero output of the ReLU neuron blocks it from any 
future gradient update. Then, we change all the activation functions from 
Leaky ReLU to ReLU while keeping the weights unchanged, retrain the network 
to recover performance, and continue to the pruning phase.

\section{Conclusions}

In this work, we combined H-LSTM and GP training to learn compact, fast, 
and accurate LSTMs. An H-LSTM adds hidden layers to control gates as opposed 
to conventional architectures that just employ a one-level nonlinearity. GP 
training combines gradient-based growth and magnitude-based pruning to ensure
H-LSTM compactness.  We GP-trained H-LSTMs for the image captioning and speech 
recognition applications. For the NeuralTalk architecture on the MSCOCO 
dataset, our models reduced the number of parameters by 38.7$\times$ (FLOPs by 
45.5$\times$) and run-time latency by 4.5$\times$, and improved the CIDEr score 
by 2.6. For the DeepSpeech2 architecture on the AN4 dataset, our models 
reduced the number of parameters by 19.4$\times$ (FLOPs by 23.5$\times$),
run-time latency by 15.7\%, and WER from 12.9\% to 8.7\%. 

\bibliographystyle{IEEEtran} 
\bibliography{bibib}

\end{document}